\documentclass[conference,10pt]{IEEEtran}
\usepackage{epsfig,rotating,setspace,latexsym,amsmath,epsf,amssymb,amsfonts,bm,theorem,subfigure,epstopdf}
\usepackage{cite,authblk}
\usepackage{bbm}
\usepackage{color}
\usepackage{mathtools}
\usepackage{algorithm}
\usepackage{algpseudocode}
\usepackage{verbatim}
\usepackage{xcolor}

\makeatletter
\newcommand\fs@betterruled{%
  \def\@fs@cfont{\bfseries}\let\@fs@capt\floatc@ruled
  \def\@fs@pre{\vspace*{5pt}\hrule height.8pt depth0pt \kern2pt}%
  \def\@fs@post{\kern2pt\hrule\relax}%
  \def\@fs@mid{\kern2pt\hrule\kern2pt}%
  \let\@fs@iftopcapt\iftrue}
\floatstyle{betterruled}
\restylefloat{algorithm}
\makeatother

\algrenewcommand\algorithmicforall{\textbf{foreach}}
\algrenewcommand\algorithmicindent{0.8em}

\algnewcommand\algorithmicforeach{\textbf{for each}}
\algdef{S}[FOR]{ForEach}[1]{\algorithmicforeach\ #1\ \algorithmicdo}

\IEEEoverridecommandlockouts
\allowdisplaybreaks

\begin{document}

\title{Transformer-Aided Semantic Communications}

\author[1]{Matin Mortaheb}
\author[1]{Erciyes Karakaya}
\author[2]{Mohammad A. (Amir) Khojastepour}
\author[1]{Sennur Ulukus}

\affil[1]{\normalsize University of Maryland, College Park, MD}
\affil[2]{\normalsize NEC Laboratories America, Princeton, NJ}

\maketitle

\begin{abstract}
The transformer structure employed in large language models (LLMs), as a specialized category of deep neural networks (DNNs) featuring attention mechanisms, stands out for their ability to identify and highlight the most relevant aspects of input data. Such a capability is particularly beneficial in addressing a variety of communication challenges, notably in the realm of semantic communication where proper encoding of the relevant data is critical especially in systems with limited bandwidth. In this work, we employ vision transformers specifically for the purpose of compression and compact representation of the input image, with the goal of preserving semantic information throughout the transmission process. Through the use of the attention mechanism inherent in transformers, we create an attention mask. This mask effectively prioritizes critical segments of images for transmission, ensuring that the reconstruction phase focuses on key objects highlighted by the mask. Our methodology significantly improves the quality of semantic communication and optimizes bandwidth usage by encoding different parts of the data in accordance with their semantic information content, thus enhancing overall efficiency. We evaluate the effectiveness of our proposed framework using the TinyImageNet dataset, focusing on both reconstruction quality and accuracy. Our evaluation results demonstrate that our framework successfully preserves semantic information, even when only a fraction of the encoded data is transmitted, according to the intended compression rates.
\end{abstract}

\section{Introduction}
In the rapidly changing realm of digital communication, efficient data transmission remains a critical challenge, particularly in situations where bandwidth is limited. This paper presents a novel approach to address this issue, utilizing transformer-aided semantic communication. A key aspect of our methodology is to tackle the bandwidth limitations posed by constrained communication channels. In numerous practical applications, bandwidth for data transmission is frequently limited, highlighting the necessity for effective data compression methods. Our approach is grounded in the principles of semantic communication, focusing not just on the transmission of raw data but more on conveying its underlying meaning or semantics. For example, in holographic telepresence \cite{holographicBo}, communication typically demands significant bandwidth to encode the data gathered from multiple sources. However, when faced with bandwidth limitations, our goal shifts towards transmitting only the crucial elements of the holographic content, such as key points, that allow the receiver to accurately reconstruct the desired aspects of the objects. This method not only preserves the integrity of the communication but also optimizes the use of available bandwidth.
 
We propose a framework to enhance the preservation of semantic information during communication processes using a vision transformer. We specifically use the attention mechanism within transformers to generate an attention mask. This mask is crucial in enhancing semantic communication by concentrating attention on significant image segments, especially those crucial for identifying key objects for inference tasks, during the reconstruction phase. Furthermore, our framework promotes efficient bandwidth utilization by encoding different parts of an image in accordance with their semantic information content. This approach can be further simplified by selectively encoding the most important parts of an image.

\begin{figure}[t]
    \centerline{\includegraphics[width=1\linewidth]{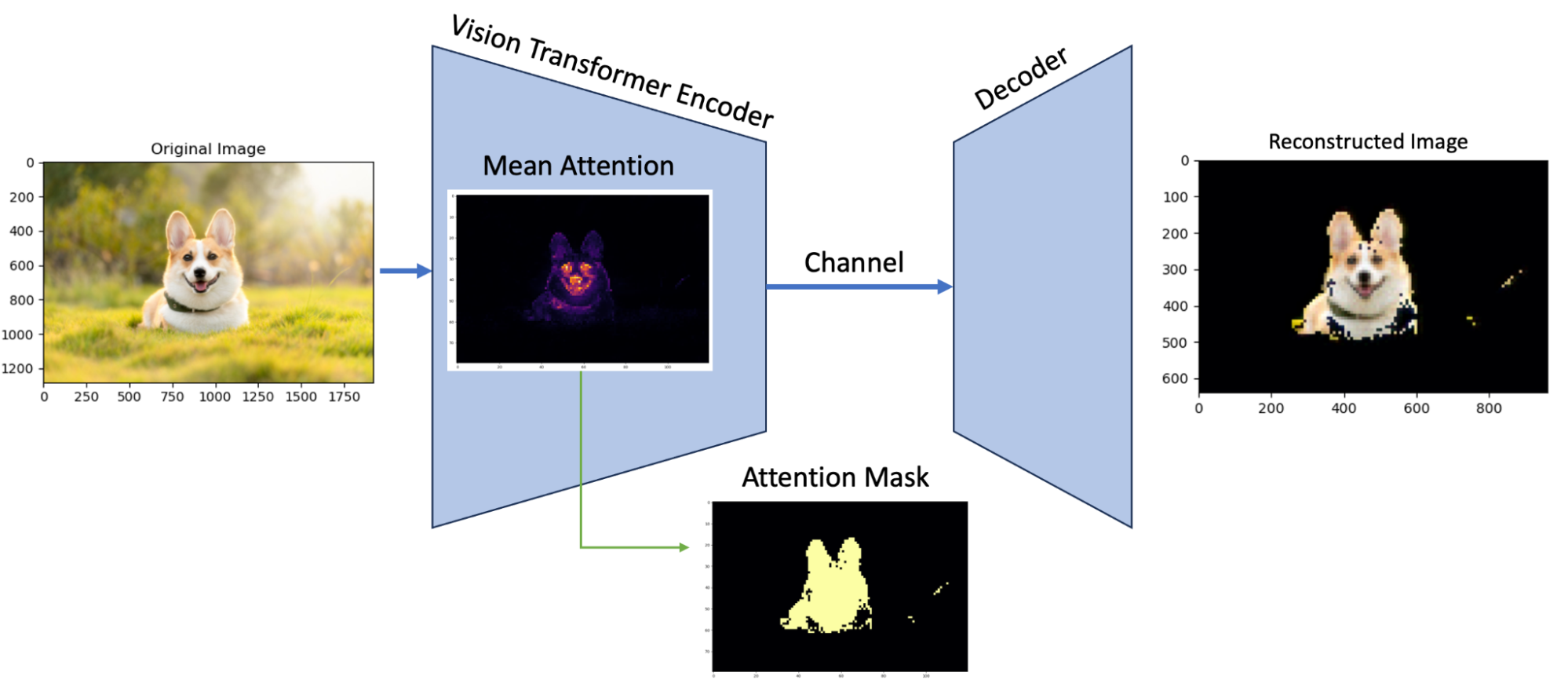}}
    \caption{Transformer-aided semantic communication framework.}
    \centering
    \label{fig:framework}
    \vspace*{-0.4cm}
\end{figure}

Several studies \cite{sagduyu2023task, sagduyu2023vulnerabilities, mortaheb2023semantic} have introduced end-to-end communication systems aimed at achieving semantic communication. These systems focus on reconstructing the image at the receiver side while ensuring the retention of semantic information during the encoding process. Typically, this approach incorporates a semantic classifier on the receiver side during training, guiding the encoded data to preserve critical elements of the image for analytical purposes. However, this method faces a limitation due to the encoder's static size, which cannot adapt to fluctuating bandwidth conditions. Moreover, encoding the entire dataset into a fixed-size format complicates the task of highlighting the most crucial segments of the original data. To overcome this challenge, the authors in \cite{reich2023deep} introduce a macroblock-wise quantization method. This approach allows for encoding the original data at the macroblock level, tailored to their significance as determined by the receiver's analysis. Nonetheless, this method necessitates an advanced deep learning (DL) module capable of deciding the optimal encoding quality for each macroblock.

\begin{figure*}[h]
    \centerline{\includegraphics[width=1\linewidth]{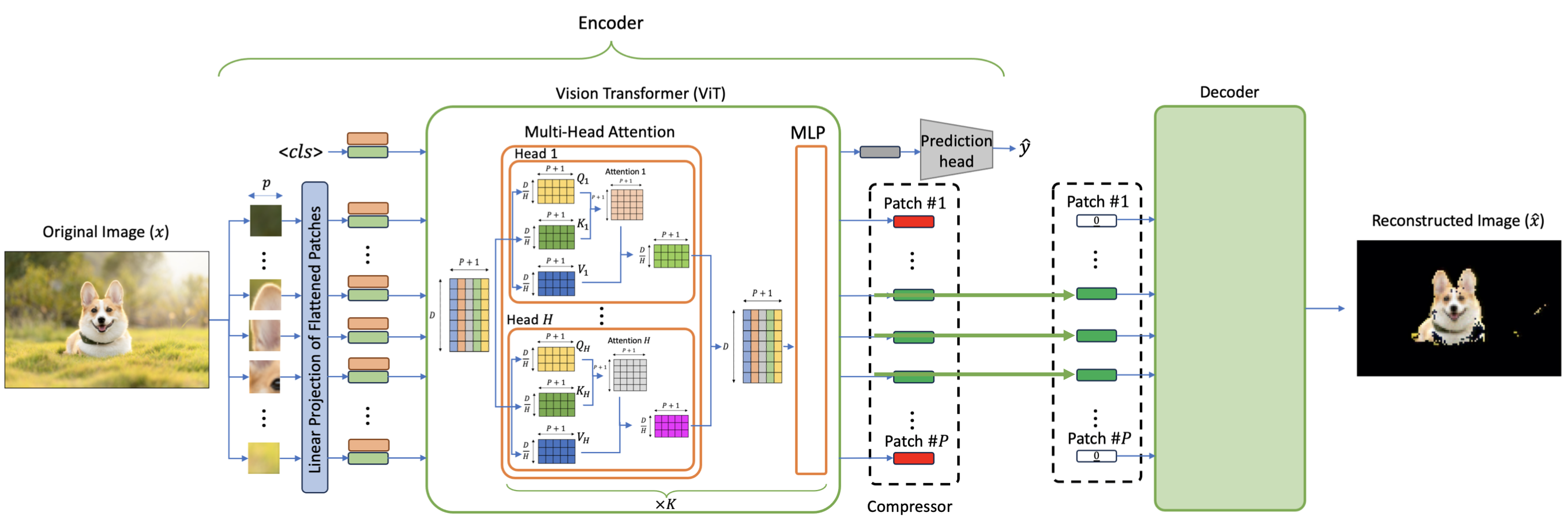}}
    \caption{System model for transformer-aided semantic communication.}
    \centering
    \label{system_model}
    \vspace*{-0.4cm}
\end{figure*}

Transformers, as introduced by \cite{vaswani2017attention}, revolutionize data encoding by leveraging the interconnections within different sections of the input. The effectiveness of transformers primarily stems from their attention unit, which assigns attention scores to various parts of the input, highlighting their relevance to the intended task. In this work, we focus on image data and utilize the vision transformer (ViT) as introduced in \cite{dosovitskiy2020image}. Previous studies \cite{yang2023witt, zhang2022unified, wu2023transformer} have explored using transformers for semantic communication. For instance, \cite{zhang2022unified} employs a transformer to create a unified model that encodes multimodal data for various tasks. \cite{wu2023transformer} introduces an end-to-end framework for wireless image transmission, utilizing transformers to harness the semantics of the source image and feedback signals received in each interaction to generate coded symbols effectively. However, to the best of our knowledge, prior studies have not explored the use of transformers to design efficient semantic communications under limited channel bandwidth constraints. 

The transformer's attention unit excels at evaluating relationships, both among the patches themselves and between each patch and the specific task the transformer is trained to perform. For example, in classification tasks, the attention unit assesses the interdependency among patches during training. Additionally, it evaluates the contribution of each patch towards accurately predicting the image's class, thereby determining the relevance of each segment in achieving the task's objective. In this paper,  we leverage the capability of the attention module within the vision transformer encoder to introduce a framework that effectively compresses encoded data, as illustrated in Fig.~\ref{fig:framework}. Our approach focuses on transmitting only the most informative patches to the receiver, which are capable of contributing meaningfully towards achieving a specific objective, such as classification. This method not only conserves bandwidth but also enhances the efficiency and effectiveness of the data analysis process on the receiver's end. We evaluate the effectiveness of our framework in terms of reconstruction and accuracy on the real-world dataset, TinyImageNet \cite{TinyIV}, across various compression rates ($r=0.25, 0.5, 0.75, 1$). The results demonstrate our framework's capacity to preserve semantic information effectively for different compression rates. 

\section{System Model and Problem Formulation} \label{sec:prob_formulation}
We formulate the problem in the context of image data. The goal is to encode an image under a bitrate constraint such that the reconstruction performance of different parts of the image is proportional to their semantic information content. Hence, the problem involves segmenting the image into patches. In this work, we consider a simple scheme where we select a subset of patches and use similar rates for encoding each of the selected patches. The selected patches correspond to the section of the image which contains more relevant information to the semantic content, e.g., the pieces of the faces of the dog in Fig.~\ref{fig:framework} where the semantic content is the class ``dog.'' We leave more sophisticated schemes, where different patches are encoded with different rates for future work. We build our solution based on a ViT. The end-to-end semantic communication system is composed of an encoder, a decoder, and a classifier where the encoder consists of three blocks: projector, ViT, and compressor. 

\subsection{Projector}
The input is an image from a set $\mathcal{X} \in \mathbb{R}^{(3, h, w)}$, where $h$ and $w$ denote the image's height and width and $3$ is the number of image channels, i.e., RGB colors. Each image $X \in \mathcal{X}$ is divided into sequences of patches $X_i$, $\forall i \in [P]$, each of size  $p \times p$, resulting in $P = \frac{hw}{p^2}$ patches per image. We also use a one-dimensional indexing (1D index) for the position of the patches by numbering the patches sequentially. Following this, each patch $X_i$ is flattened to vector $\bar{X}_i \in \mathbb{R}^{(3p^2,1)}$ which are in turn transformed into vectors $\Acute{X}_i$ of dimension $D$ via $W_{proj}\in\mathbb{R}^{(D, 3p^2)}$ that is analogous to tokenization in natural language processing (NLP) tasks. 

\subsection{ViT}
For each image, ViT first forms the matrix $\Tilde{X} = [\Tilde{X}_{cls},\Tilde{X}_1,\cdots,\Tilde{X}_P] \in \mathbb{R}^{(D, P+1)}$ which consist of a random vector $\Tilde{X}_{cls}$ and sequence of $P$ vectors $\Tilde{X}_i$ for each patch in order, where $\Tilde{X}_i$ is the superposition of $\Acute{X}_i$ with the positional encoding vectors for the respective patches. After passing through ViT, the transformation of the vector $\Tilde{X}_{cls}$ will hold the concise information for classification of the image. 

The matrix $\Tilde{X}$ undergoes processing through $K$ transformer layers sequentially, each of which containing multi-head attention (MHA) unit followed by a multi-layer perceptron (MLP) layer $f_{MLP}$ with a non-linear activation function. The output of the last transformer layer  $z=[z_{cls},z_1,\cdots,z_P] \in \mathbb{R}^{(D,P+1)}$ comprises $z_{cls}$ which is the transformation of $\Tilde{X}_{cls}$ used for classification, followed by $P$ vectors $z_i$ that contain the encoded data for patch $i$.

Each transformer layer in ViT consists of $H$ parallel attention heads. For simplicity of the notation, we drop the indexing of the layers and in the following we explain a generic layer. The input $\Tilde{X}$ is vertically partitioned into $H$ matrices, i.e., $\Tilde{X}=[\Tilde{X}_1,\cdots,\Tilde{X}_H]^T$ where $\Tilde{X}_h \in \mathbb{R}^{(D/H,P+1)}, \forall h \in [H]$ is the input for the head $h$. 

The query $Q^{(h)}$, key $K^{(h)}$, and value $V^{(h)}$ matrices of size $D/H$ by $P+1$ are produced by using the trainable set of weight matrices $W_Q^{(h)}$, $W_K^{(h)}$, and $W_V^{(h)}$, of size $D/H$ by $D/H$, respectively. For each head $h$, the attention score matrix $A^{(h)}(Q,K) \in \mathbb{R}^{(P+1, P+1)}$ for head $h$ is then calculated as follows,
\begin{align}
\text{A}^{(h)}(Q, K) = \text{softmax}\left(\frac{Q^{(h)}K^{{(h)}^T}}{\sqrt{D}}\right)
\end{align} 

The output for head $h$ is obtained by a linear transformation of the value matrix $V^{(h)}$ through attention scores $A^{(h)}$. The output from each attention head $O^{(h)}$ is concatenated to form the output of the MHA unit, represented as $O = [O^{(1)},\cdots,O^{(H)}] \in \mathbb{R}^{(D, P+1)}$. This output $O$ is then fed into an MLP with transfer function $f_{MLP}(O, \phi_{mlp})$ to produce the output for the transformer layer, where $\phi_{mlp}$ is the MLP parameters. Note that each layer has its own parameters that are learned in the training phase.

ViT predicts the input image class $\hat{y}$ by processing $z_{cls}$ through a single-layer neural network $f_{predictor}$, i.e., $\hat{y}=f_{predictor}(z_{cls},\phi_{pred})$, where $\phi_{pred}$ is the parameters of the prediction head.

\subsection{Compressor, Decoder, and Classifier}
The compressor prepares transmission packets based on the output from the ViT and its internal states. We assume that the channel is error-free and it has limited capacity which may fluctuate over time. The compressor's role is to adapt the packet bitrates to match the instantaneous channel capacity. The received packets are re-arranged in a proper format and passed to the decoder, which reconstructs the image. The classifier processes the reconstructed image to predict its class.

\section{Main Contributions}
We introduce the use of transformers for the purpose of semantic communications by compressing the output of the transformer blocks of a ViT based on the internal state of the ViT, namely, the attention score matrix. Specifically, we select a portion of the patches based on their attention scores such that the rate of the transmitted packet for the selected image is within the instantaneous bitrate constraint of the channel. The proposed scheme is built on using the attention score of a patch as a representative of its information metric with respect to the semantic content.

First, we perform supervised training of ViT by jointly training the projector, ViT, and its dedicated prediction head $f_{predictor}$.
During the training phase, we employ cross-entropy loss $\mathcal{L}_{CE}(\hat{y},c)$ between the true class label $c$ and $\hat{y}$ to jointly train the projector, transformer, and classifier.

\begin{figure}[t]
    \centerline{\includegraphics[width=1\linewidth]{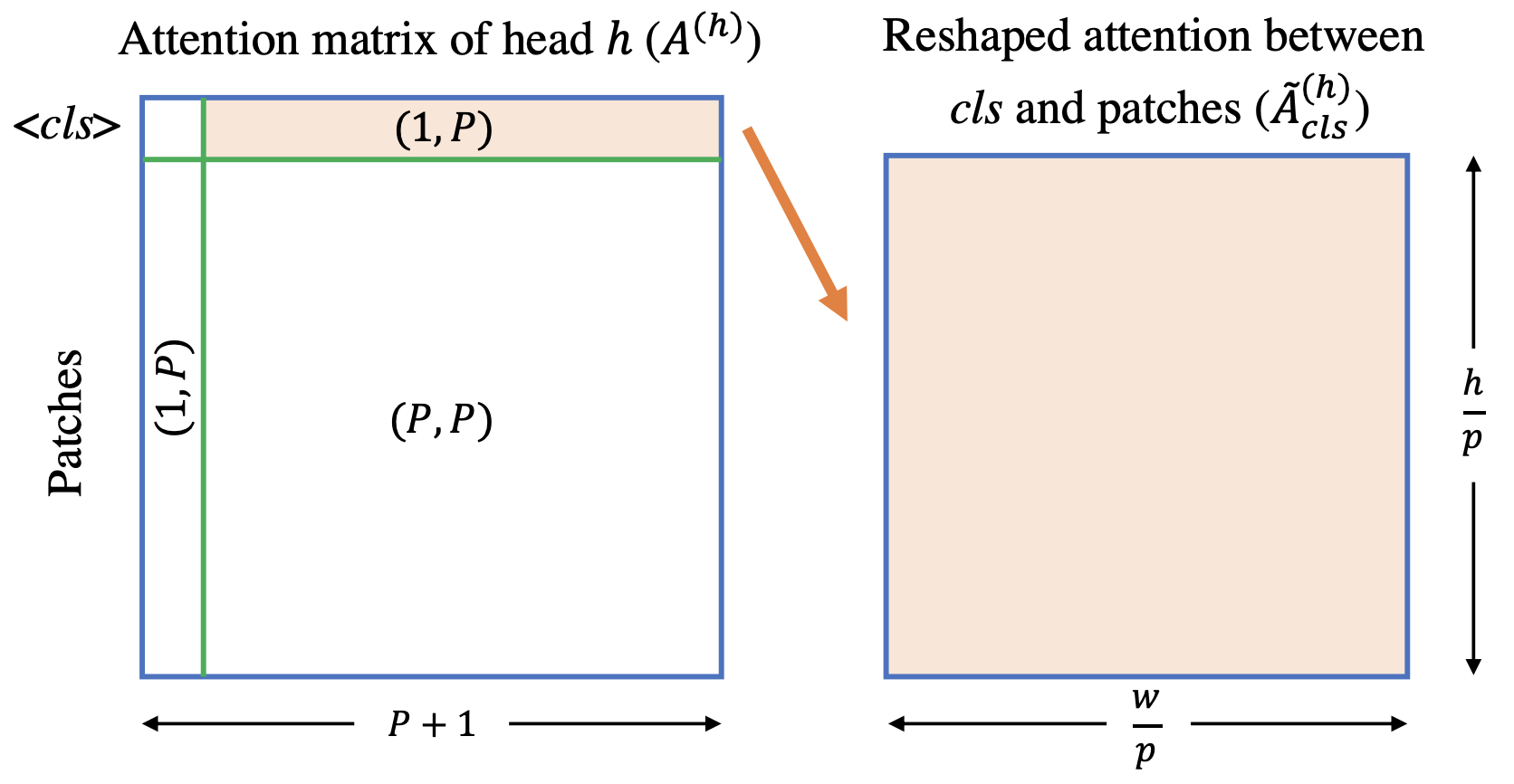}}
    \caption{Deriving $\Tilde{A}_{cls}^{(h)}$ from $A^{(h)}$.}
    \centering
    \label{fig:reshape_attn}
    \vspace*{-0.4cm}
\end{figure}

\begin{figure}[t]
    \centerline{\includegraphics[width=1\linewidth]{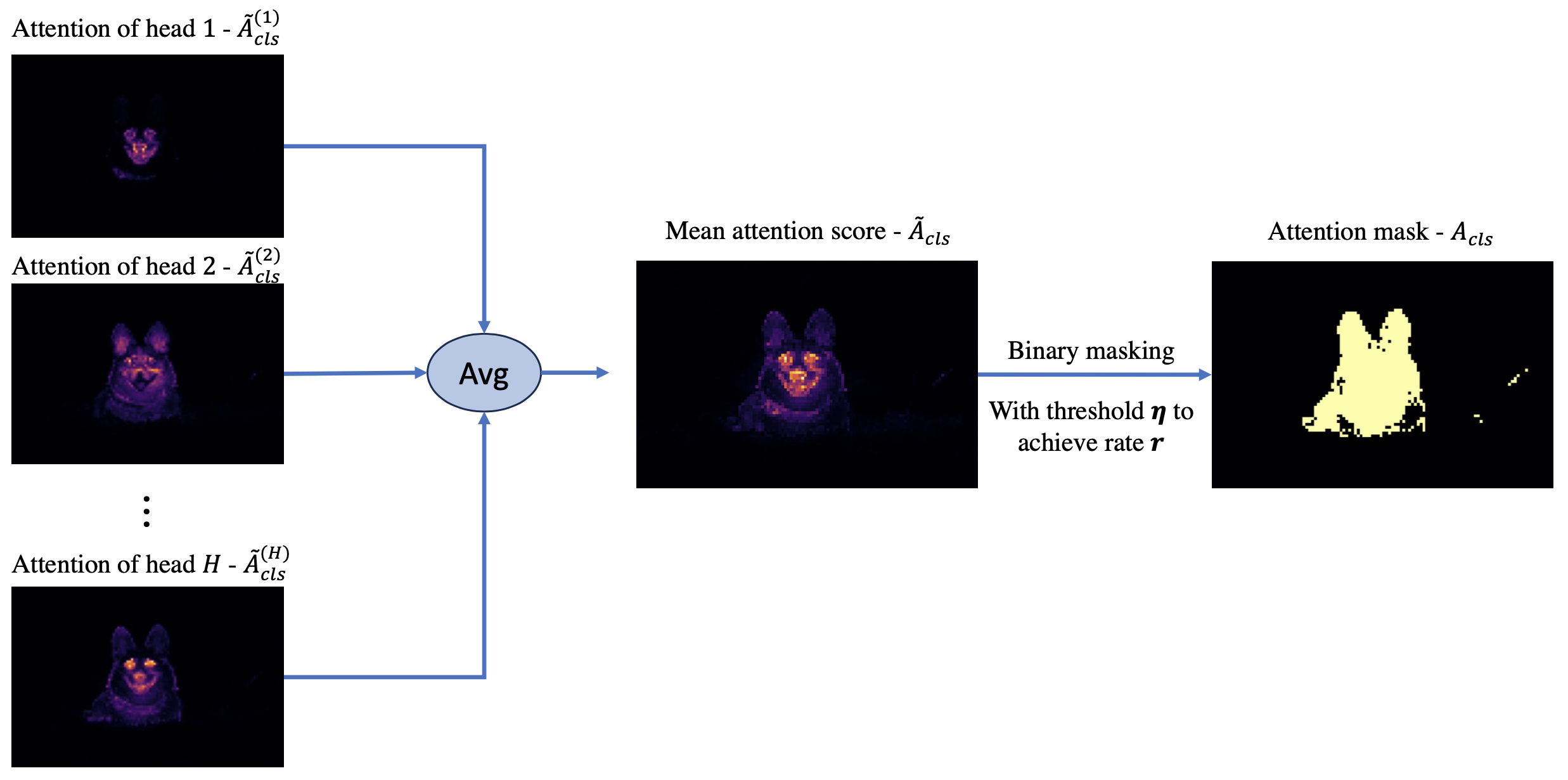}}
    \caption{Obtaining attention mask from ViT attention unit.}
    \centering
    \label{fig:attention_mask}
\end{figure}

For a trained transformer the attention score matrices $\{A^{(h)}, h \in [H]\}$ in the last transformer layer contain information about the semantic content of the patches. Illustrated in Fig.~\ref{fig:reshape_attn}, we use the first row of the attention matrix denoted by $A_{cls}^{(h)} = A^{(h)}[0,:], h \in [H]$ as a measure of the relevancy of patches to the semantic content of the input image. More precisely, the value of the score in position $i$ in vector $A_{cls}^{(h)}$ highlights the significance of the patch with 1D index $i$ for accurate classification. It is convenient to reshape $A_{cls}^{(h)}$ into a square matrix of size $(w/p, h/p)$ to form $\Tilde{A}_{cls}^{(h)}$ by using the relationship between the two-dimensional position and 1D index of the patches. In order to combine the information from all heads, we calculate the average of $\Tilde{A}_{cls}^{(h)}$ for all $h \in [H]$ heads denoted as $\Tilde{A}_{cls}$. Fig.~\ref{fig:attention_mask} shows this averaging process and highlights its effect in recognizing the position of the patches that are important in an illustrative example. 

We aim to use these scores to create a binary mask, allowing only those patches that positively affect the classification to be transmitted as marked by one as opposed to zero for the patched that are not transmitted. 

The compressor uses a design parameter $\alpha \in [0,1]$ to find a threshold $\lambda$ such that by selecting the patches with higher scores than $\lambda$ their cumulative bitrate is maximized but does not exceed $\alpha$ portion of the bitrate constraint. Next, the compressor selects additional random patches from the remaining ones to maximize the transmission rate without exceeding the bitrate constraint. Hence, the transmitted packet $\hat{z}$ consists of the encoded data for the selected patches, and a positional mask $M \in \mathbb{R}^{(w/p, h/p)}$ which indicates which patches are selected. In our simulations, we have ignored the bitrate required for the positional mask as it is negligible in comparison to the packet bitrate.  

The decoder $f_{decoder}(\hat{z},\theta)$ is designed to reconstruct the image by minimizing the loss function 
\begin{align} \label{eq: loss_dec}
    \mathcal{L}_{decoder} = ||(M \otimes B) \odot (X- \hat{X})||^2
\end{align}
by training the parameters $\theta$, which is aimed to minimize the mean squared error (MSE) for the selected patches in $M$ where $B$ is a matrix of size $p \times p$, $\odot$ is the Hadamard product, and $\otimes$ is the Kronecker product. Based on our assumption of attention score as the representative of the information metric with respect to the semantic content, this design choice can be interpreted as maximizing the reconstruction performance of different parts of the image proportional to their semantic information. The overall algorithm is shown in Algorithm~\ref{alg:first_method}.

\begin{algorithm}[!ht]
    \caption{Training with our proposed algorithm}
    \label{alg:first_method}
    \begin{algorithmic}[1]
    \State {\bfseries Input:} {Image $X \in \mathbb{R}^{(3, h, w)}$, $c$: class of the image}
    \State {\bfseries Initialization:} {Embedding dimension: $D$, path size: $p$, number of heads: $H$}
    \State {\bfseries Parameters:} Transformer encoder parameters at time $t$: $\Theta_{t}$:=$\{$\small{$\left\{W_Q^{(h)},
    W_K^{(h)}, W_V^{(h)}\right\}_{h=1}^H$\normalsize, $\phi_{mlp}$, $\phi_{pred},W_{proj}\}$
    \Statex Decoder parameters at time $t$: $\theta_t$}

    \Statex \hrulefill
    \State {Segmenting $X$ into $P$ patches each with dimension of $(3, p, p)$, where 3 stands for RGB data}
    \State {\bfseries Do} $\bar{X}_i\leftarrow$flatten($X_i$), $\bar{X}_i\in\mathbb{R}^{(3p^2,1)}$, $\forall i \in [P]$
    \State Define $W_{proj}\in\mathbb{R}^{(D,3p^2)}$ with bias $b$ \small\Comment{Linear Projection}\normalsize
    \State {\bfseries Do} $\Acute{X}_i\leftarrow(W_{proj}\cdot\bar{X}_i + b)^T$, $\widehat{X}_i\in\mathbb{R}^{(D,1)}$, $\forall i \in [P]$
    \State Add positional encoding to $\Acute{X}_i$ to form $\Tilde{X}_i$
    \State $\Tilde{X}=[\Tilde{X}_{cls}, \Tilde{X}_1, ..,\Tilde{X}_P]$ s.t. $\Tilde{X}\in \mathbb{R}^{(D,P+1)}$ \Comment{Enc. input}
    \State Partition $\Tilde{X}$ into $H$ chunks s.t. $\Tilde{X}=[\Tilde{X}_1; \Tilde{X}_2;\cdots;\Tilde{X}_{H}]$
    \Statex where $\Tilde{X}_h\in \mathbb{R}^{(\frac{D}{H},P+1)}, \forall h \in [H]$
    \While {$t$ $<$ end time}\Comment{Encoder training}
    \ForEach {$h \in [H] $ (in parallel)}\Comment{MHA}
    \small
    \State $Q^{(h)}\leftarrow W_Q^{(h)}\cdot\Tilde{X}_h$, $K^{(h)}\leftarrow W_K^{(h)}\cdot\Tilde{X}_h$, $V^{(h)}\leftarrow W_V^{(h)}\cdot\Tilde{X}_h$
    \normalsize
    \State $\text{A}^{(h)}\leftarrow \text{softmax}\left(\frac{Q^{(h)}\cdot{K^{(h)}}^T} {\sqrt{D}}\right)$ \Comment{Attention Score}
    \State $\text{O}^{(h)}\leftarrow {V^{(h)}}^T\cdot\text{A}^{(h)}$
    \EndFor
    \State MHA output: $O=[O^{(1)},\cdots,O^{(H)}]\in\mathbb{R}^{(D, P+1)}$
    \State Assign $z\leftarrow f_{MLP}(O,\phi_{mlp})$ \Comment{Encoded data $z$}
    \Statex \hspace*{0.15cm} Recall that $z=[z_{cls},z_1, z_2,\cdots, z_P] \in \mathbb{R}^{(D,P+1)}$
    \State Determine the class $\hat{y}=f_{predictor}(z_{cls},\phi_{pred})$
    \State Calculate $\mathcal{L}_{encoder}=\mathcal{L}_{CE}(\hat{y},c)$
    \State $\Theta_{t+1}=\Theta_{t}-\eta_1\nabla\mathcal{L}_{encoder}$
    \EndWhile
    \While {$t$ $<$ end time}\Comment{Decoder training}
    \ForEach {$h \in [H] $ (in parallel)}
    \State $\Tilde{A}_{cls}^{(h)} = A^{(h)}[0,:]$
    \State Reshape $\Tilde{A}_{cls}^{(h)}$: $\mathbb{R}^{(1, P)} \longrightarrow \mathbb{R}^{(\frac{h}{p},\frac{w}{p})}$
    \EndFor
    \State $\Tilde{A}_{cls}=\mathbb{E}[\Tilde{A}_{cls}^{(h)}]$
    \For {$\forall i \in [\frac{h}{p}]$ and $\forall j \in [\frac{w}{p}]$} \Comment{Attention mask}
    \State \textbf{if} {$\Tilde{A}_{cls}[i][j]\leq\lambda$} \textbf{then} $A_{cls}[i][j]\leftarrow$0\textbf{else} $A_{cls}[i][j]\leftarrow1$
    \EndFor
    \State Reshape $A_{cls}^{(h)}$: $\mathbb{R}^{(\frac{h}{p},\frac{w}{p})} \longrightarrow \mathbb{R}^{(P)}$
    \For {$\forall i \in [P]$} \Comment{Transmitted encoded data}
    \State \textbf{if} {$A_{cls}[i]==0$} \textbf{then} $\hat{z}[i]$ = 0, \textbf{else} $\hat{z}[i] = z[i]$
    \EndFor
    \State $\hat{X}=f_{decoder}(\hat{z},\theta_t)$ \Comment{Reconstruct the image $\hat{x}$}
    \State Expand $A_{cls}$ to $(h,w)$ dimensions
    \State $\mathcal{R}(X)\leftarrow X\cdot A^T, \mathcal{R}(\hat{X})\leftarrow \hat{X}\cdot A^T$
    \State Calculate $\mathcal{L}_{decoder}=||(M \otimes B) \odot (X- \hat{X})||^2$
    \State $\theta_{t+1}=\theta_{t}-\eta_2\nabla\mathcal{L}_{decoder}$
    \EndWhile
    \end{algorithmic}
\end{algorithm}

\begin{figure}[t]
    \centerline{\includegraphics[width=0.65\linewidth]{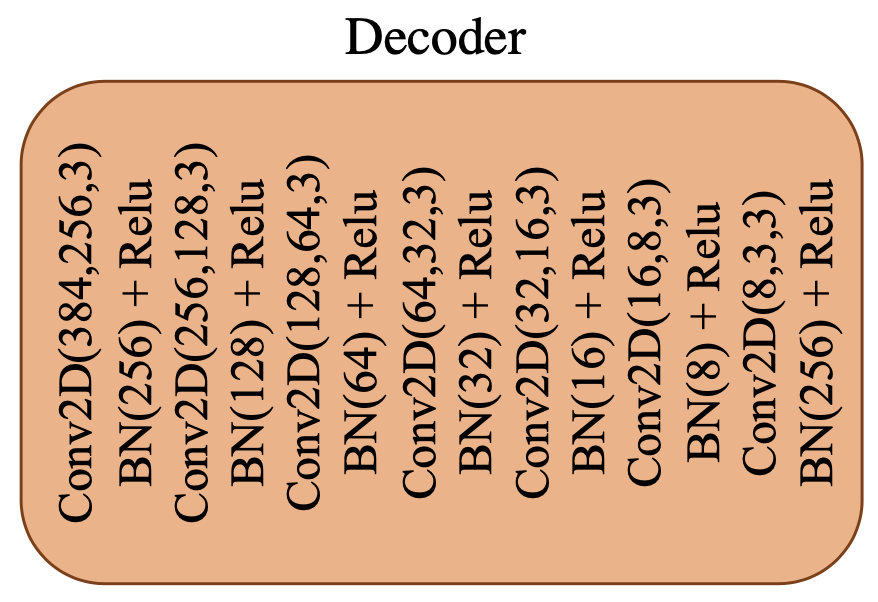}}
    \caption{Decoder model structure.}
    \centering
    \label{fig:decoder_model}
    \vspace*{-0.4cm}
\end{figure}

\section{Experimental Results} \label{sec:Exp_section}

\subsection{Dataset Specifications}
We use TinyImageNet dataset \cite{TinyIV} to fine-tune both encoder and decoder parameters in our framework. TinyImageNet is a scaled-down version of the larger ImageNet dataset \cite{imagenet}, containing fewer images and categories. Specifically, TinyImageNet has 200 classes, each with 500 training images, 50 validation images, and 50 test images, making it a total of 100,000 training images, 10,000 validation images, and 10,000 test images. All images in the TinyImageNet dataset are originally 64x64 in RGB format. For our purposes, we scale-up these images to a resolution of $480\times320$. The transformer encoder structure we use has been previously trained on the larger ImageNet dataset \cite{imagenet}.

\subsection{Model and Hyperparameters}
Our proposed framework consists of two primary components: the encoder and the decoder. In the encoder part, we utilize the DINO \cite{DINO} model as a transformer encoder, which is pretrained through a self-supervised approach on the ImageNet dataset. DINO contains three variants: \emph{Tiny-ViT}, \emph{Small-ViT}, and \emph{ViT}. For our framework, we select \emph{Small-ViT} with a patch size of $p = 8$. This model has an embedding dimension of $D = 384$, and includes $K = 12$ transformer blocks, each with $H = 6$ attention heads. On the decoder side, we have designed a deep neural network (DNN) model that features a sequence of convolutional layers. The detailed structure of this decoder model is illustrated in Fig.~\ref{fig:decoder_model}. We use Adam optimizer with learning rate $\eta = 5\times10^{-4}$ to individually train the model parameters. The batch size is set to 32. To calculate the accuracy result, we use a pre-trained classifier trained on TinyImageNet dataset.

\subsection{Results and Analysis}
\begin{figure}[t]
    \centerline{\includegraphics[width=1\linewidth]{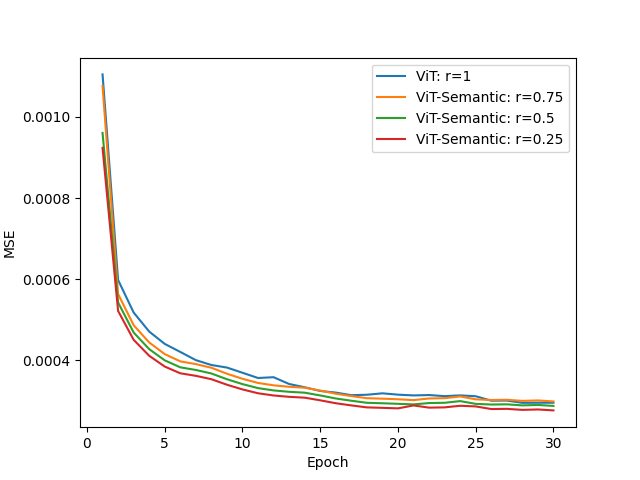}}
    \caption{MSE performance between the compressed and uncompressed model in our framework.}
    \centering
    \label{fig:mse_performance}
    \vspace*{-0.4cm}
\end{figure}

We evaluate the performance of our proposed method in terms of reconstruction and accuracy performance. We conduct the training at 4 different rates: $r=1, 0.75, 0.5, 0.25$. For rate $r=1$, we send the entire encoded data to the decoder side. Therefore, it can be considered as a traditional semantic communication model which does not use the attention module to compress the encoded data. For rate $r=0.5$, we use the same framework but we only send 50\% of the entire patches based on our derived attention mask. To fairly compare the reconstruction performance on all the rates, we calculate the MSE between the original and reconstructed images on the selected patches which are flagged by a 1 in the attention mask \ref{eq: loss_dec}. We show the reconstruction performance  in Fig.~\ref{fig:mse_performance}. The reconstruction performance indicates that the compressed case can be reconstructed as good as (or slightly better) than the uncompressed version on the area requested from the attention mask to be reconstructed.

\begin{figure}[t]
    \centerline{\includegraphics[width=0.8\linewidth]{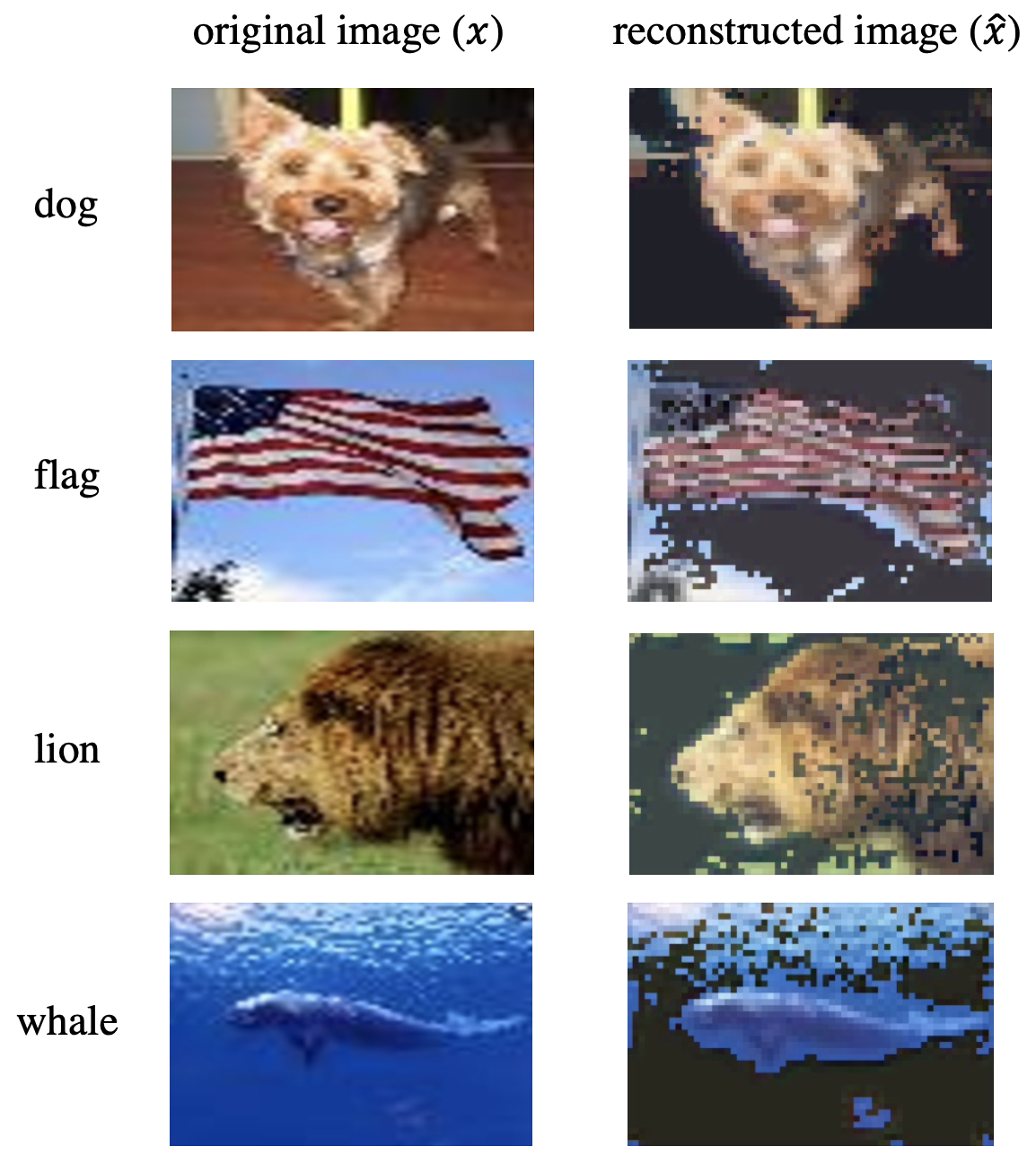}}
    \caption{Example of reconstructed image with compression of $r=0.5$.}
    \centering
    \label{fig:recon_example}
    \vspace*{-0.4cm}
\end{figure}

Fig.~\ref{fig:recon_example} demonstrates the original and reconstructed images for some instances of the test data. We observe that the transformer encoder is able to effectively identify and extract the most informative parts of the data, including the essential information needed for classification purposes. 

\begin{figure}[t]
    \centerline{\includegraphics[width=0.75\linewidth]{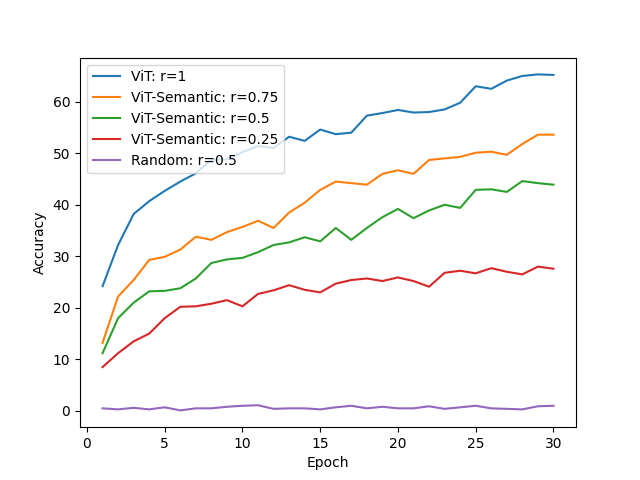}}
    \caption{Accuracy for $\alpha = 1$.}
    \centering
    \label{fig:acc_performance}
    \vspace*{-0.4cm}
\end{figure}

To evaluate the accuracy of the reconstructed images within our framework, we employ a classifier that was trained on the original TinyImageNet dataset. We examine the accuracy across all four compression rates by using a compressor with $\alpha = 1$, as shown in Fig.~\ref{fig:acc_performance}. The findings confirm that our framework is capable of maintaining the semantic integrity of the images, even when reconstructing only a small portion of the patches. This effectiveness is attributed to the capability of the model to preserve the most critical parts of the image because of the attention mask. 

\begin{figure}[t]
    \centerline{\includegraphics[width=0.75\linewidth]{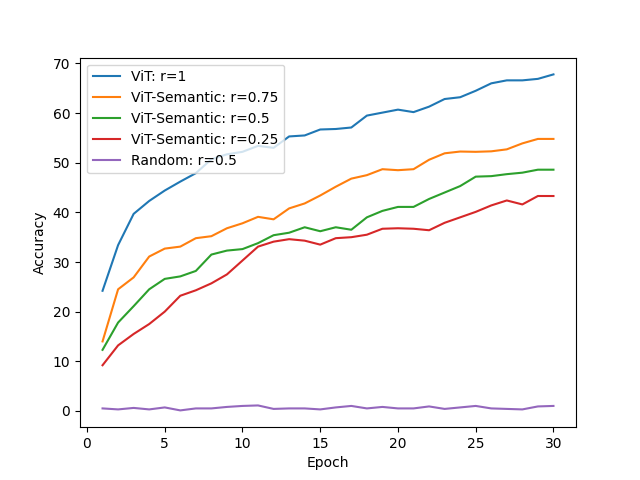}}
    \caption{Accuracy for $\alpha = 0.85$.}
    \centering
    \label{fig:acc_performance_add}
    \vspace*{-0.4cm}
\end{figure}

Next, we examine the effect of the parameter $\alpha$ used by the compressor in selecting the patches. Fig.~\ref{fig:acc_performance_add} illustrates the effect of using $\alpha= 0.85$ in comparison to the result shown in Fig.~\ref{fig:acc_performance}. We see that by lowering $\alpha$ by 15\% the accuracy is improved, in particular, for lower compression rates. For example, the accuracy is improved from 24\% to 43\% over 30 epochs for the compression rate $r = 0.25$. This can be explained by noting that we use the attention score merely as representative of the information in a patch with respect to the semantic content, and even though we believe there is a strong correlation between these two, the true relationship has not been established. Even if the attention score and the semantic information content are almost linearly connected, we are only allowed to remove a patch completely if its semantic information content is zero, otherwise, it cannot be completely ignored and the compressor should still allocate a partial (i.e., non-zero) bitrate to such patches. Since we use a simple binary mask, we seek the remedy by selecting a random number of chunks with lower attention scores (i.e., below the threshold $\lambda$). 

\begin{figure}[t]
    \centerline{\includegraphics[width=0.75\linewidth]{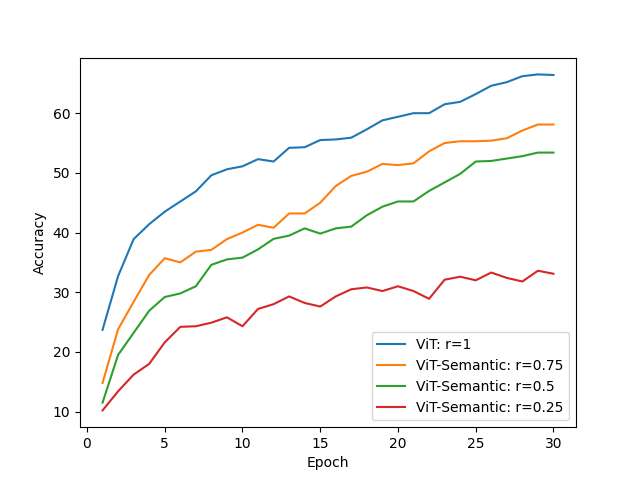}}
    \caption{Accuracy using a fine-tuned classifier.}
    \centering
    \label{fig:acc_performance_fined}
    \vspace*{-0.4cm}
\end{figure}

Finally, we note that the classifier which is designed to work on the original images may not be optimized for the classification of the reconstructed images based on the binary mask as we have proposed in this work since as part of the image is going to be missing. The missing (or blackened) background in the reconstructed images can act as an adversarial input for the classifier since the classifier is only seen the original images with backgrounds. Therefore, in order to improve the classification accuracy, we propose to fine-tune the classifier with the following loss function 
\begin{align}
    \mathcal{L}_{classifier} = \beta \mathcal{L}_{CE}(y(X),c) + (1-\beta) \mathcal{L}_{CE}(y(\hat{X}),c)
\end{align}
where $y(\cdot)$ denotes the pre-trained classifier that we use for TinyImageNet dataset. The loss incorporates the cross entropy for the reconstructed images based on the compressed data (defined as \emph{compressed images}) in addition to the cross-entropy of the original images (i.e., \emph{uncompressed images}) into the loss function. Fig.~\ref{fig:acc_performance_fined} shows the classification accuracy performance for different compression rates using the fined-tuned classifier using the above loss function with $\beta = 0.3$. 

As shown in Fig.~\ref{fig:acc_performance_fined}, there is a 10\% improvement in accuracy for $r=0.5$, and approximately, a 6\% improvement for both $r=0.25$ and $r=0.75$. 

\section{Discussion}
Semantic communication frameworks typically perform simultaneous training of both encoder and decoder parameters based on a loss function that combines MSE and cross-entropy $\mathcal{L}_{CE}$ losses, e.g.,
\begin{align} \label{eq:loss}
    \mathcal{L} & = \gamma \overbrace{||x -\hat{x}||^2}^\text{MSE loss} + (1-\gamma) \overbrace{\mathcal{L}_{CE}(\hat{y},c)}^\text{Semantic Loss}  
\end{align}
which involves both analytic accuracy of semantic content and reconstruction performance of the original data \cite{mortaheb2023semantic, sagduyu2023task}. However, in our approach, we first train the encoder using the cross-entropy loss and subsequently train the decoder based on the MSE loss due to the following three reasons/observations: (i) The convergence is slow when using joint training of encoder and decoder with combined loss function. (ii) Even though the encoder is trained only based on the cross-entropy loss and without directly considering the MSE loss, the decoder that is trained subsequently is able to reconstruct the patches based on the encoder output with negligible MSE. (iii) Besides the compressor and its parameters, the design of the other part of the encoder, i.e., the projector and ViT can be performed independent of the channel quality, e.g., bitrate constraint. 

\bibliographystyle{unsrt}
\bibliography{reference}
\end{document}